\documentclass[conference]{IEEEtran}
\IEEEoverridecommandlockouts
\usepackage{cite}
\usepackage{amsmath,amssymb,amsfonts}
\usepackage{algorithmic}
\usepackage{textcomp}
\usepackage{xcolor}
\usepackage{array}
\usepackage{makecell}
\usepackage[ruled,vlined,linesnumbered]{algorithm2e}
\usepackage{subcaption} 
\usepackage{adjustbox}

\usepackage{url}
\usepackage{latexsym}
\usepackage{multirow}
\usepackage{footnote}
\usepackage{CJKutf8}
\usepackage{booktabs}
\usepackage{microtype}
\usepackage[font=normal]{caption}

\usepackage{listings}
\usepackage{spverbatim}

\definecolor{codegreen}{rgb}{0,0.6,0}
\definecolor{codegray}{rgb}{0.5,0.5,0.5}
\definecolor{codepurple}{rgb}{0.58,0,0.82}
\definecolor{backcolour}{rgb}{0.95,0.95,0.92}

\lstdefinestyle{mystyle}{
    backgroundcolor=\color{backcolour},   
    commentstyle=\color{codegreen},
    keywordstyle=\color{magenta},
    numberstyle=\tiny\color{codegray},
    stringstyle=\color{codepurple},
    basicstyle=\ttfamily\footnotesize,
    breakatwhitespace=false,         
    breaklines=true,                 
    captionpos=b,                    
    keepspaces=false,                 
    numbers=left,                    
    numbersep=5pt,                  
    showspaces=false,                
    showstringspaces=false,
    showtabs=false,                  
    tabsize=2
}

\usepackage[font=normal]{caption}
   
\usepackage[most]{tcolorbox}
\definecolor{block-gray}{gray}{0.85}
\newtcolorbox{myquote}{colback=block-gray,grow to right by=-10mm,grow to left by=-10mm,
boxrule=0pt,boxsep=0pt,breakable}

\def\BibTeX{{\rm B\kern-.05em{\sc i\kern-.025em b}\kern-.08em
    T\kern-.1667em\lower.7ex\hbox{E}\kern-.125emX}}
\begin{document}

\title{DataVisT5: A Pre-trained Language Model for Jointly Understanding Text and Data Visualization}

\author{
    \IEEEauthorblockN{Zhuoyue Wan\IEEEauthorrefmark{1}, Yuanfeng Song\IEEEauthorrefmark{2}, Shuaimin Li\IEEEauthorrefmark{1}, Chen Jason Zhang\IEEEauthorrefmark{1}, Raymond Chi-Wing Wong\IEEEauthorrefmark{3}}
    \IEEEauthorblockA{\IEEEauthorrefmark{1}PolyU, Hong Kong, China 
    \IEEEauthorrefmark{2}WeBank Co., Ltd, Shenzhen, China 
    \IEEEauthorrefmark{3}HKUST, Hong Kong, China
    }
}

\maketitle

\begin{abstract}
Data visualization (DV) is the fundamental and premise tool to improve the efficiency in conveying the insights behind the big data, which has been widely accepted in existing data-driven world. Task automation in DV, such as converting natural language queries to visualizations (i.e., text-to-vis), generating explanations from visualizations (i.e., vis-to-text), answering DV-related questions in free form (i.e. FeVisQA), and explicating tabular data (i.e., table-to-text), is vital for advancing the field. 
Despite their potential, the application of pre-trained language models (PLMs) like T5 and BERT in DV has been limited by high costs and challenges in handling cross-modal information, leading to few studies on PLMs for DV. We introduce \textbf{DataVisT5}, a novel PLM tailored for DV that enhances the T5 architecture through a hybrid objective pre-training and multi-task fine-tuning strategy, integrating text and DV datasets to effectively interpret cross-modal semantics. Extensive evaluations on public datasets show that DataVisT5 consistently outperforms current state-of-the-art models and higher-parameter Large Language Models (LLMs) on various DV-related tasks. We anticipate that DataVisT5 will not only inspire further research on vertical PLMs but also expand the range of applications for PLMs.


\end{abstract}

\begin{IEEEkeywords}
pre-trained language model, data visualization, text-to-vis, vis-to-text, FeVisQA, table-to-text
\end{IEEEkeywords}

\section{Introduction}
Data visualizations (DVs) utilize graphical representation to convey insights to summarize the massive raw data, which is a common practice in existing big data era \cite{friendly2008brief,qin2020making}. 
Popular data analysis and database applications, such as Google Sheets\footnote{\url{https://www.google.com/sheets/about/}} and Microsoft Power BI\footnote{\url{https://powerbi.microsoft.com/}}, all support DV features.
Many institutions realize the value of DV and have applied it as their daily fundamental tools. Thus the ability of creating suitable DVs has become a necessary skill for data analysts, engineers, and data scientists \cite{stodder2013data,knaflic2015storytelling,zheng2017data}. 
However, creating appropriate DVs remains challenging, even for experts, since it requires visual analysis expertise and familiarity with the domain data. Furthermore, users must master the complex grammar of \emph{Declarative Visualization Languages} (DVLs), such as Vega-Lite \cite{satyanarayan2016vega}, ggplot2 \cite{villanueva2019ggplot2}, and Vega-Zero \cite{luo2021natural}, to accurately define DV \emph{specification} in the visualization engine.

\begin{figure*}[t!]
    \centering
    \includegraphics[width=1.0\linewidth]{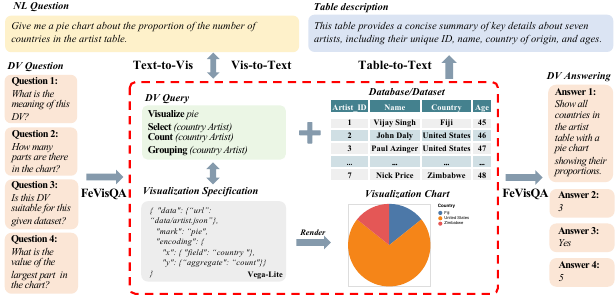}
    \vspace{-10pt}
    \caption{An illustration depicting the \textit{text-to-vis}, \textit{vis-to-text}, \textit{table-to-text}, and \textit{free-form question-answering over data visualization} problems, showcasing examples including a NL question, a DV query, a DVL visualization specification, a table description, a visualization chart, and four question-answer pairs.}
    \label{fig:example}
    \vspace{-15pt}
\end{figure*}

To lower the barriers to creating DVs and further unlock the power of DV for the general public, researchers have proposed a variety of DV-related tasks that have attracted significant attention from both industrial and academic researchers. Numerous studies on these topics have been presented in leading conferences and journals such as VLDB \cite{vartak2015seedb,siddiqui2016effortless,qin2020making}, ICDE \cite{luo2018deepeye2,song2024}, SIGMOD \cite{hanrahan2006vizql,luo2018deepeye,luo2021synthesizing}, and TKDE \cite{luo2020steerable,zhang2024_dvsurvey}. These tasks include \emph{text-to-vis} (i.e., automatically generating DVs from natural language questions) \cite{luo2021natural,luo2021synthesizing}, \emph{vis-to-text} \cite{song2020vis2text} (i.e., automatically generating interpretations of complex DVs for educational purposes), \emph{FeVisQA} \cite{song2024} (i.e., free-form question answering over data visualization), and \emph{table-to-text} (i.e., describing a given table) \cite{kasner-etal-2023-tabgenie}.

A vivid example is given in Figure~\ref{fig:example}, which shows four important tasks central to the domain knowledge of DV: \emph{text-to-vis}, \emph{vis-to-text}, \emph{FeVisQA} 
and \emph{table-to-text}. The figure presents a natural language (NL) question, \emph{``Give me a pie chart about the proportion of the number of countries in the artist table.''} This example demonstrates the text-to-vis task's capability to interpret the NL question and transform it into a Vega-Lite specification, resulting in a pie chart. The \emph{DV query}, introduced by \cite{luo2021synthesizing}, serves as a bridge in the text-to-vis process, encapsulating visualization details and data operations with a grammar akin to SQL. Translations between DV queries and DVLs are seamless, with text-to-vis tasks primarily focusing on converting NL questions into DV queries. 
Conversely, the vis-to-text task aims to generate accessible and user-friendly explanations of complex visualizations for individuals without expertise in the field. The FeVisQA task addresses user inquiries regarding DV by providing detailed answers to common questions. We present four typical DV-related questions, including understanding the semantics of a DV query, resolving numerical issues within a chart, and evaluating the compatibility of a DV query with a given database. Lastly, the table-to-text task generates informative NL description of tabular data, which are essential for visual analytics, thereby reducing the perceptual effort needed for data interpretation.

Meanwhile, PLMs such as BERT \cite{devlin2019bert} and T5 \cite{raffel2019exploring} have received considerable attention in the realms of natural language processing (NLP) and data mining, becoming widely recognized for their efficacy.
These PLMs greatly promote the development of effective text-driven applications, since they show dominating performance in understanding the semantics in natural language. 
The operational paradigm for these PLMs typically unfolds in two stages: initially, they undergo unsupervised pre-training on expansive, open-domain datasets (such as Wikipedia) to acquire foundational capabilities in language representation and comprehension; subsequently, they are fine-tuned on specialized corpora pertinent to targeted downstream tasks, thereby enhancing task-specific performance.
Despite their success \cite{wang2021codet5,edwards-etal-2022-translation,UnifiedSKG}, there are still significant challenges when it comes to the DV field : (i) Limited studies have been conducted to explore the effectiveness of PLMs in capturing the DV semantics.  (ii) Since there is a substantial modal gap between the DV modality and the text modality, satisfied performances cannot be achieved by directly applying existing PLMs (e.g., T5) to DV-related tasks mentioned above. (iii) In the DV area, a possible PLM needs the ability of handling cross-modal information (i.e., text and DV), while also being capable of managing multiple distinct tasks.

To alleviate above-mentioned problems, we propose a novel PLM for jointly understanding text and DV, refereed as \textbf{DataVisT5} in this paper. Based on text-centric T5 architecture,  we enhance the pre-training process by incorporating a comprehensive array of cross-modal datasets that integrate natural language with DV knowledge, encompassing DV queries, database schemas, and tables. 
Since DV queries resemble programming language-like queries, we employ CodeT5+ \cite{wang2023codet5p} as the starting checkpoint in our work. This choice leverages the robust code semantic understanding and generation capabilities of CodeT5+, providing DataVisT5 with a substantial advantage in generating and comprehending the unique programming language of our DV tasks.
Building on this foundation,  we apply \textit{table-level database schema filtration} to reduce training complexity. Addressing the challenges of format consistency between DV and textual modalities, we introduce a \textit{unified encoding format} for DV knowledge that facilitates the convergence of text and DV modalities. To eliminate stylistic discrepancies in manually curated datasets, we adopt standardized encoding.

Additionally, the pre-training objectives for DataVisT5 are twofold: (i) the span corruption approach of Masked Language Modeling as utilized by the original T5 model, and (ii) a Bidirectional Dual-Corpus objective that operates on source-target pairings.
After the mixed-objective pre-training, we conduct multi-task fine-tuning (MFT) of our DataVisT5 on DV-related tasks including text-to-vis, vis-to-text, FeVisQA, and table-to-text.
To substantiate the rationale behind our proposed model, we performed comprehensive experimental evaluations on various public datasets. The results consistently demonstrate that DataVisT5 surpasses the state-of-the-art (SOTA) models and higher-parameter LLMs.
In summary, our main contributions are as follows:
\begin{itemize}
\setlength{\parsep}{0pt}
\setlength{\parskip}{0pt}
\item We introduce and release DataVisT5: the first Pre-trained Language Model (PLM) tailored for the joint understanding of text and DV. This innovation opens avenues for future research on task-specific PLMs and enriches the landscape of PLM designs.
\item We enhance the text-centric T5 architecture to handle cross-modal information. Our novel hybrid pre-training objectives are conceived to unravel the complex interplay between DV and textual data, fostering a deeper integration of cross-modal insights.
\item Extensive experiments on public datasets for diverse DV tasks including text-to-vis, vis-to-text, FeVisQA, and table-to-text demonstrate that DataVisT5 excels in multi-task settings, consistently outperforming strong baselines and establishing new SOTA performances.

\end{itemize}

\section{Preliminary}
\label{sec:pre}
This section provides the foundational concepts and definitions pivotal to DV-related tasks, with the objective of cultivating a more profound understanding.

\noindent\textbf{Natural Language Question.} An NL question enables users, even those with a minimal background in DV and programming skills, to formulate queries intuitively. 
Figure~\ref{fig:example} demonstrates such an instance, with the user's request articulated as, ``\emph{Give me a pie chart about the proportion of the number of countries in the artist table}''.

\noindent\textbf{Declarative Visualization Language.} Transforming data into a graphical representation typically involves the use of a declarative visualization language (DVL). This kind of language provides a set of specifications that determine the construction of visualizations. These specifications include various elements such as chart type, colors, sizes, and mapping functions, as well as properties for visual marks like canvas dimensions and legends. Several DVLs are prevalent in the field, such as Vega-Lite \cite{satyanarayan2016vega}, ggplot2 \cite{villanueva2019ggplot2}, ZQL \cite{siddiqui2016effortless}, ECharts \cite{li2018echarts}, Vega-Zero \cite{luo2021natural}, and VizQL \cite{hanrahan2006vizql}, each offering unique features to facilitate the visualization process.

\noindent\textbf{Visualization Specification.} A visualization specification comprises a \emph{JSON} format object that delineates the dataset and its visual attributes (such as chart types and data transformation functions) in accordance with the syntax of a specific DVL. It is noteworthy that each DVL possesses a unique grammar, necessitating distinct visualization specifications for rendering the same DV chart.

\noindent\textbf{Data Visualization Query.}
Introduced by \cite{luo2018deepeye, luo2018deepeye2}, a framework for querying a database for visual data representations seeks to encapsulate the full spectrum of potential DVLs. 
As depicted in Figure~\ref{fig:example}, a DV query specifies a "pie" chart and integrates SQL-like operations (e.g. \emph{Count} and \emph{Order By}). This versatile DV query format can be converted into visualization specifications for different DVLs, enabling visualization engines to render the specified chart.

\noindent\textbf{Data Visualization Chart.} 
The DV charts are the visual representations such as scatters, bars, or maps used to convey the data summary and insights defined by the visualization specification. In Figure~\ref{fig:example}, the final visualization result is the bar chart that corresponds to the NL question.


\section{Our Proposed Model: DataVisT5}
\label{sec:model}

We present our proposed DataVisT5 model, with the pipeline overview in Section~\ref{subsec:overview}. This is followed by details on database schema filtration in Section~\ref{subsec:dbfilter}, DV knowledge encoding in Section~\ref{subsec:dv_enc}, and standardized encoding in Section~\ref{subsec:standard}. We discuss our hybrid pre-training objectives in Section~\ref{subsec:preobj} and conclude with our multi-task fine-tuning strategy in Section~\ref{subsec:MTF}.

\begin{figure}[t!]
    \centering
\includegraphics[width=0.48\textwidth]{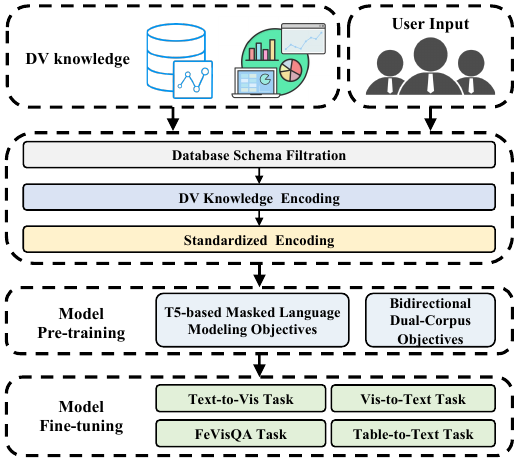}
\vspace{-5pt}
    \caption{The pipeline of DataVisT5.}
\label{fig:dvt5}
\vspace{-15pt}
\end{figure}

\subsection{Pipeline Overview}
\label{subsec:overview}

Figure~\ref{fig:dvt5} provides an overview of the complete pipeline, comprising five main stages: (1) \emph{Database schema filtration}, (2) \emph{DV knowledge Encoding}, (3) \emph{Standardized Encoding}, (4)\emph{Model Pre-training}, and (5) \emph{Model Fine-tuning}. 
%
The \emph{Database schema filtration} process involves comparing n-grams extracted from the given database schema with those present in the text under consideration, enabling us to identify referenced tables in the question and acquire a sub-database schema that aligns semantically. During the \emph{DV knowledge Encoding} phase, we linearize DV knowledge encompassing DV queries, database schemas, and tables. Subsequently, in the \emph{Standardized Encoding} phase, we normalize the DV knowledge to facilitate more efficient learning. The resulting corpus, in its unified form, is then employed to train our proposed DataVisT5 model.


\vspace{-2.5pt}
\subsection{Database Schema Filtration}
\label{subsec:dbfilter}
Before the integration of DV and text modalities, it is critical to recognize that NL questions can incorporate keywords related to the database schema. This requires the explicit identification of references to columns, tables, and conditional values within the NL questions. To address this challenge, we employ $N$-gram matching as a method due to its simplicity of implementation and notable effectiveness for a variety of applications. 
In an effort to minimize information loss, our primary focus is at the table level, where we compare $N$-grams extracted from the NL questions to those present in the database tables. Following the initial comparison, we refine the obtained sub-schema by considering the implicated tables and their respective columns.



\vspace{-2.5pt}
\subsection{DV Knowledge Encoding}
\label{subsec:dv_enc}
To address the disparity between text and DV modalities, we propose investigating unified formats for DV knowledge. The connection between natural language and DV knowledge poses challenges due to limited data accessibility. Nevertheless, a unified format allows models to capitalize on extensive pretraining for smaller datasets. Employing consistent formatting, as recommended by \cite{giaquinto-etal-2023-multitask}, offers advantages in multi-task training and mitigates performance decline caused by data heterogeneity compared to single-task training. The subsequent sections provide a comprehensive introduction to the unified representation of three distinct types of DV knowledge: DV queries, database schemas, and tables. 

\begin{figure}[t!]
    \centering
    \includegraphics[width=0.48\textwidth]{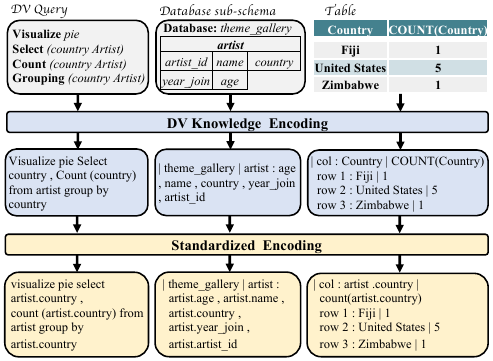}
    \caption{Examples of DV Knowledge Encoding and Standardized Encoding from NVBench.}
    \label{fig:dvstd}
    \vspace{-15pt}
\end{figure}

\noindent\textbf{Encoding DV query.}
While most existing NLP models, such as \cite{devlin2019bert}, consider NL inputs as flat text sequences, we adopt a similar approach for modeling a DV query by treating it as a plain text sequence in a straightforward manner.

\noindent\textbf{Encoding Database schema.}
The database schema comprises tables and columns. For each table in the schema, the table name is followed by a list of its columns formatted as "$table:$ $column_1$, ... $column_n$". Different tables are joined using the symbol "$|$". Additionally, the database name is prefixed to the generated sequence with boundaries indicated by "$|$".

\noindent\textbf{Encoding Table.}
Following \cite{herzig2020tapas}, we employ a sequential representation of tables, akin to the schema encoding technique, which uses distinctive tokens to delineate table structure. The table is linearly represented as ``$col: c_1 \mid \cdots \mid c_N$ $row\ 1: v_{11} \mid \cdots \mid v_{1N}$ $\cdots$ $row\ M: v_{M1} \mid \cdots \mid v_{MN}$'', with \(N\) indicating the total column count and \(M\) representing the row count.

\noindent\textbf{Example.} 
An presented in Figure~\ref{fig:dvstd}, where (1) the DV query is sequentially encoded into text sequences based on the data manipulation operations: \emph{Visualize}, \emph{Select}, \emph{Count}, and \emph{Grouping}, (2) the filtered database sub-schema, including the database name (\emph{theme\_gallery}), table name (\emph{artist}), and columns, is encoded into a corresponding text sequence, and (3) the table content is linearly encoded in the format ``\emph{col: Country $\mid$ COUNT(Country)}'', along with the remaining three rows of the table. 

\subsection{Standardized Encoding}
\label{subsec:standard}
Due to the manual generation of queries by multiple annotators with diverse annotation habits, subtle stylistic differences are prevalent in the final annotated DV queries within NVbench, including variations in the capitalization of keywords. Similar to issues encountered with SQL queries, these stylistic inconsistencies, while not affecting the model's execution results, pose an additional learning challenge that must be addressed. To address the stylistic variations in DV queries, a preprocessing strategy was implemented before training. This strategy includes: (1) affixing the primary table name \textit{T} to the selected columns \textit{col}, resulting in the notation \textit{T.col} across DV queries; particularly, for instances where the wildcard symbol \textit{*} is employed in a \textit{COUNT} function, \textit{COUNT(*)} is replaced with \textit{COUNT(T.col)} to maintain uniformity; (2) the insertion of spaces surrounding parentheses and the replacement of double quotes with single quotes; (3) the inclusion of the \textit{ASC} keyword subsequent to the \textit{ORDER\ BY} clause when ordering is not explicitly specified; (4) the elimination of the \textit{AS} clause and the substitution of table aliases (e.g., \textit{T1}, \textit{T2}) with their actual table names; (5) the lowercase conversion.

\begin{figure}[th!]
    \centering    \includegraphics[width=0.48\textwidth]{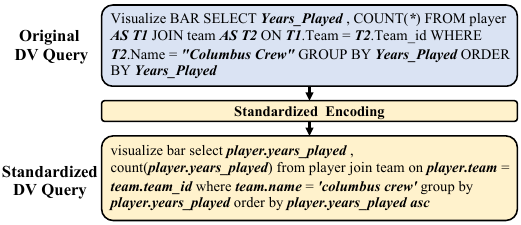}
    \caption{An Standardized DV Query with join operation example.}
    \label{fig:dvq_eg}
    \vspace{-10pt}
\end{figure}

\noindent\textbf{Example.} 
In a DV query with a $Join$ operation, as depicted in Figure~\ref{fig:dvq_eg}, standardization involves renaming table aliases $T1$ and $T2$ to $player$ and $team$, respectively. The query's $COUNT(*)$ is specified as $COUNT(player.years\_{played})$, 'Columbus Crew' is quoted with single quotes, the $ASC$ keyword is appended if sort order is absent, and the entire query is cast to lowercase.


In alignment with the standardization of DV queries, similar encoding steps are applied to database schemas and tables to ensure consistency. This includes affixing the table name $T$ to each column name $col$ and converting them to $T.col$.

\begin{figure*}[ht!]
    \centering
    \includegraphics[width=1\textwidth]{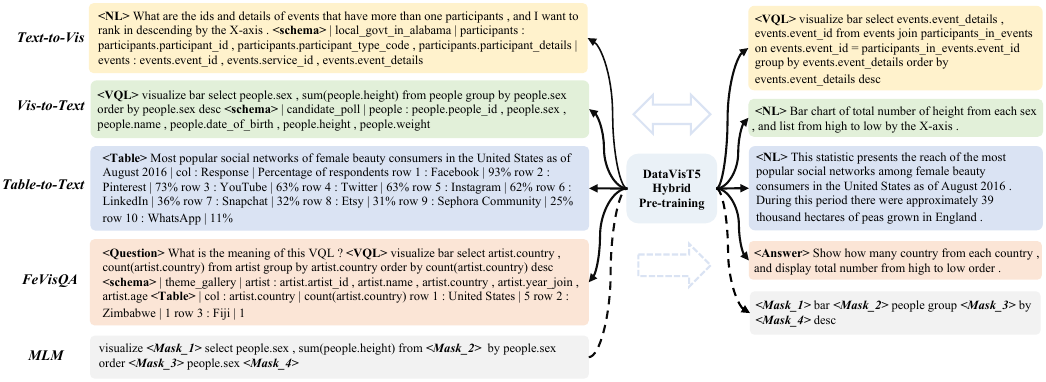}
    \vspace{-15pt}
    \caption{Overview of hybrid pre-training objectives. The solid lines denote the Bidirectional Dual-Corpus objectives, which facilitate the learning of language representation by leveraging bidirectional context. The dashed lines represent the T5-based MLM objectives, designed to reconstruct the original input from masked tokens.}
    \label{fig:pre_frame}
    \vspace{-15pt}
\end{figure*}

\noindent\textbf{Example.} 
As depicted in Figure~\ref{fig:dvstd}, within a specific database schema, column names such as "\emph{age}, \emph{name}, \emph{country}, \emph{year\_join}, and \emph{artist\_id}" are transformed to "\emph{artist.age}, \emph{artist.name}, \emph{artist.country}, \emph{artist.year\_join}, and \emph{artist.artist\_id}", respectively. Similarly, within the table context, an entry like "\emph{col : Country $\mid$ COUNT(Country)}" is reformulated to "\emph{col : artist.country $\mid$ count(artist.country)}".

\subsection{Hybrid Pre-training Objectives}
\label{subsec:preobj}
\noindent\textbf{Bidirectional Dual-Corpus Objectives.} To address divergence between the pretraining and fine-tuning phases, we introduce Bidirectional Dual-Corpus (BDC) objectives during pretraining.In this approach, both the source and target corpora are randomly selected with equal probability (0.5) during model training to serve as the input. The remaining corpus is then used as the output for translation purposes.
Accordingly, for a target sequence of $T$ tokens, we define the BDC loss function, $\mathcal{L}_{BDC}(\theta)$, as follows:
\begin{equation}
\mathcal{L}_{BDC}(\theta) = \sum_{i=1}^{T} -\log P_{\theta}(t_{i} \mid \mathbf{s}, \mathbf{t}_{<i}),
\end{equation}
where $\mathbf{s}$ signifies the source input, $\mathbf{t}_{<i}$ represents the sequence of tokens generated by the decoder up to but not including the $i$-th token, and $t_{i}$ is the token that the decoder is tasked with predicting. The term $\theta$ denotes the model parameters.

As depicted in Figure~\ref{fig:pre_frame}, the segment highlighted by arrows elucidates the deployment of the BDC Objectives, encompassing four discrete tasks germane to DV. A comprehensive definition of these tasks is deferred to Section~\ref{sec:exp}.
To enhance task-specific processing and facilitate knowledge transfer across different modalities, we introduce unique special tokens.
For example, as demonstrated in Figure~\ref{fig:pre_frame}, the Text-to-Vis task utilizes a special token \textless$NL$\textgreater\ to prefix the NL question corpus and \textless$VQL$\textgreater\ for the DV query corpus. In contrast, for the FeVisQA task, DV question-answer pairings are delineated with the tokens \textless$Question$\textgreater\ and \textless$Answer$\textgreater\ to signify their respective components.

\noindent\textbf{T5-based MLM Objectives.}
The application of Masked Language Modeling (MLM) as a pretraining objective is pivotal for pretraining encoder-decoder models. In our study, we employed the span corruption MLM strategy from \cite{raffel2019exploring}, where consecutive words in the input are replaced by sentinel tokens, and the decoder generates the omitted text, each instance preceded by its respective sentinel. To ensure consistency with the pretraining checkpoint, we maintained an average span length of 3 subword tokens across the input corpus and masked 15\% of the subwords. This MLM objective was applied to a cross-modal corpus comprising text, DV query, database schema, and table.
Over a sequence of $N$ tokens, our T5-based MLM loss is defined as:
\begin{equation}
\mathcal{L}_{MLM}(\theta) = \sum_{n=1}^{N} -\log P_{\theta}\left(x_{n}^{\text{m }} \mid \mathbf{x}^{\backslash \text{m }}, \mathbf{x}_{<n}^{\text{m }}\right),
\end{equation}
where $\theta$ are the model parameters, $x_{n}^{\text{m}}$ is the masked token predicted by the decoder, $\mathbf{x}^{\backslash \text{m}}$ represents the unmasked encoded inputs, and $\mathbf{x}_{<n}^{\text{m}}$ is the sequence of tokens generated by the decoder up to but not including the $n$-th token.

An illustration is presented in Figure~\ref{fig:pre_frame}, where the segments linked by dashed lines pertain to the T5-based MLM Objectives. This figure showcases the application of span denoising targets to a DV query. Within this query, the terms "bar", "people group", "by", and "desc" are selected at random. Subsequently, a subset of these terms is replaced by sentinel tokens, illustrated as \textless Mask\_1\textgreater, \textless Mask\_2\textgreater, \textless Mask\_3\textgreater, and \textless Mask\_4\textgreater.



\noindent\textbf{Hybrid Objectives.}
After achieving the aforementioned two objectives, we create a hybrid objective by sampling from both the MLM Objectives and the BDC Objectives corpora.  Consequently, each training mini-batch is composed of examples drawn from a cross-modal corpus, each formatted to align with diverse learning objectives.
We adopt a final hybrid loss $\mathcal{L}_{H}$:
\begin{equation}
    \mathcal{L}_{H}(\theta)=\mathcal{L}_{BDC}(\theta)+\mathcal{L}_{MLM}(\theta),
\end{equation}
which enables DataVisT5's readiness for multiple DV-related downstream tasks demanding contextual comprehension and pattern recognition.

\subsection{Multi-Task Fine-tuning}
\label{subsec:MTF}
To achieve better performance in multiple downstream tasks related to DataVisT5, we employ temperature mixing to combine the training data of all tasks. The temperature value is set to 2, following \cite{raffel2019exploring}. Temperature up-sampling helps balance the influence of each task on the model by adjusting the probability of selecting data from each task during training. This prevents larger datasets from overpowering smaller ones. By merging training data from different tasks, the model is encouraged to learn representations that are beneficial across various corpora. Consequently, this leads to improved generalization and a more robust model capable of handling diverse DV tasks.

\section{Pretraining Dataset Construction}
\label{sec:data}
We have constructed a dataset tailored for our Hybrid Pretraining Objectives by integrating four public datasets. The following sections outline our pretraining dataset construction, detailing data collection in Section~\ref{subsec:data_collect}, data processing in Section~\ref{subsec:data_processing}, and data partitioning in Section~\ref{subsec:data_partition}.
\subsection{Data Collection} 
\label{subsec:data_collect}
\subsubsection{NVBench}
\label{subsubsec:nvbench}
The NVBench dataset \cite{luo2021synthesizing} represents a publicly accessible NL2Vis corpus, containing 7,219 pairs of NL questions and their corresponding DV queries. It was originally curated to evaluate the efficacy of models in transforming textual queries into visual representations. As the most commonly utilized dataset in this domain, NVBench has been employed in several prominent studies, including those by\cite{luo2021natural,song2020vis2text,10.1145/3534678.3539330}
%
Table~\ref{tab:NVBench} offers a detailed overview of the NVBench dataset, comprising 25,628 entries that have been collated from 152 distinct databases originating from the Spider dataset \cite{yu2018spider}. To facilitate fair comparison with other established baselines as discussed in Section~\ref{sec:exp}, we meticulously separated the DV queries involving non-join operations from those that include join operations and performed an in-depth statistical analysis. Specifically, the dataset contains 15,764 samples without join operations. DV queries that employ non-join operations, utilizing a single table, are showcased in Figure~\ref{fig:dvstd}. Conversely, DV queries featuring join operations, where multiple tables are engaged, are illustrated in Figure~\ref{fig:dvq_eg}.



\begin{table}[t!]
\center
\caption{The statistics of the {NVBench} dataset}
\label{tab:NVBench}
\resizebox{1\columnwidth}{!}{%
\begin{tabular}{ccccc}
\hline
 & \multicolumn{2}{c}{\textbf{Number of instances}} & \multicolumn{2}{c}{\textbf{Number of databases}} \\
\textbf{Split} & NVBench w/o join & \multicolumn{1}{c|}{NVBench} & NVBench w/o join & NVBench \\ \hline
Train & 10564 & \multicolumn{1}{c|}{16780} & 98 & 106 \\
Valid & 2539 & \multicolumn{1}{c|}{3505} & 15 & 16 \\
Test & 2661 & \multicolumn{1}{c|}{5343} & 27 & 30 \\
Total & 15764 & \multicolumn{1}{c|}{25628} & 140 & 152 \\ \hline
\end{tabular}
}
\vspace{-10pt}
\end{table}

\subsubsection{Chart2text.}
\label{subsubsec:chart2text}
The chart-to-text conversion process, as introduced by \cite{kantharaj-etal-2022-chart}, constitutes a comprehensive benchmark incorporating two distinct datasets, cumulatively consisting of 44,096 charts that span an extensive array of subjects and graphical representations. The data for this benchmark originates from two primary sources: Statista\footnote{\url{https://www.statista.com/}} and the Pew Research Center\footnote{\url{https://www.pewresearch.org/}}. The dataset derived from Statista includes various elements such as a screenshot of the chart image, the accompanying data table, the title, axis labels, and expertly crafted descriptive narratives concerning the chart content. 
Conversely, the datasets sourced from the Pew Research Center typically lack the provision of underlying data tables for the majority of their charts. To align with our pre-training objectives, we have selectively utilized only the Statista component of the Chart2Text dataset. The quantitative details of the Chart2Text dataset are systematically tabulated in Table~\ref{tab:tabledataset}, with a total of 34,811 instances documented for analysis.

\subsubsection{WikiTableText.}
\label{subsubsec:wikitabletext}
The WikiTableText dataset \cite{bao2018table} consists of 13,318 descriptive sentences that are aligned with 4,962 tables extracted from Wikipedia\footnote{\url{https://www.wikipedia.org/}}. These tables were retrieved via web scraping techniques and a subset of 5,000 tables was carefully curated to ensure that each table contained at least three rows and two columns, thereby meeting a predefined structural criterion. Quantitative characteristics of the WikiTableText dataset are meticulously cataloged in Table~\ref{tab:tabledataset}, which enumerates a total of 13,318 instances for subsequent analysis.

\begin{table}[t!]
\center
\caption{The statistics of the Chart2text and WikiTableText datasets}
\label{tab:tabledataset}
\resizebox{1\columnwidth}{!}{%
\begin{tabular}{cccccc}
\hline
 & \multicolumn{2}{c}{\textbf{Number of instances}} &  & \multicolumn{2}{c}{\textbf{Number of cells}} \\
\textbf{Split} & Chart2Text & \multicolumn{1}{c|}{WikiTableText} & \textbf{Metrics} & Chart2Text & WikiTableText \\ \hline
Train & 24368 & \multicolumn{1}{c|}{10000} & Min. & 4 & 27 \\
Valid & 5222 & \multicolumn{1}{c|}{1318} & Max. & 8000 & 108 \\
Test & 5221 & \multicolumn{1}{c|}{2000} & $\leq$150 & 34272 & 13318 \\
Total & 34811 & \multicolumn{1}{c|}{13318} & $>$150 & 539 & 0 \\ \hline
\end{tabular}%
}
\vspace{-10pt}
\end{table}

\begin{table}[t!]
\center
\caption{The statistics of the FeVisQA dataset}
\label{tab:FeVisQA}
\resizebox{1\columnwidth}{!}{%
\begin{tabular}{ccccccc}
\hline
\multicolumn{1}{l}{} & \multicolumn{3}{c}{\textbf{Number of instances}} & \multicolumn{3}{c}{\textbf{Number of questions}} \\
\textbf{Spilt} & databases & QA pair & \multicolumn{1}{c|}{DV query} & Type 1 & Type 2 & Type 3 \\ \hline
Train & 106 & 54406 & \multicolumn{1}{c|}{9169} & 4799 & 9166 & 31272 \\
Valid & 16 & 9290 & \multicolumn{1}{c|}{1603} & 844 & 1579 & 5264 \\
Test & 30 & 15609 & \multicolumn{1}{c|}{2542} & 1453 & 2501 & 9113 \\
Total & 152 & 79305 & \multicolumn{1}{c|}{13313} & 7096 & 13246 & 45650 \\ \hline
\end{tabular}%
}
\vspace{-10pt}
\end{table}

\subsubsection{FeVisQA}
\label{subsubsec:fevisqa}
The FeVisQA dataset, as presented in \cite{song2024}, represents a pivotal asset in the nascent field of DV Question Answering. This dataset amalgamates a diverse set of rules and data sources to compile a comprehensive collection of question-and-answer pairs, integral for advancing research in this domain. It covers three principal types of questions:

\begin{itemize}

\item \textit{Type 1}: This question type probes the semantic interpretation of DVs. An example is, "What is the meaning of this DV ?" which is illustrated as Question 1 in Figure~\ref{fig:example}.
\item \textit{Type 2}: Stemming from the associated task of DV recommendation, this category includes questions that assess the suitability of a DV for a given dataset. For instance, "Is this DV suitable for the given dataset?" The answers are structured to affirm compatibility or denote incompatibility, thus evaluating the alignment between a DV and its corresponding dataset.
\item \textit{Type 3}: Questions pertaining to data retrieval and the structural aspects of DV. These are generated using a rule-based approach, ensuring a robust and consistent set of questions and answers. Question 3 and Question 4 in Figure~\ref{fig:example} serve as exemplary instances of this category.
\end{itemize}
Comprehensive statistics of the FeVisQA dataset are encapsulated in Table~\ref{tab:FeVisQA}. Similar to NVBench, the FeVisQA leverages the 152 databases originating from the Spider dataset \cite{yu2018spider}, comprising a total of 79,305 free-form question-answer pairs.

\subsection{Data Pre-processing}
\label{subsec:data_processing}
To enhance the data quality and ensure compatibility with downstream tasks, we instituted the following pre-processing. Initially, we excluded incomplete natural language question samples (34/25662) from the NVBench dataset.
Subsequently, to prevent sequence truncation during the Bidirectional Dual-Corpus objective—which operates with a fixed token length—we retained only those entries in the Chart2Text dataset where the total number of cells (determined by multiplying the number of rows by the number of columns) did not exceed 150. This step was deemed unnecessary for the WikiTableText dataset, as it inherently possesses a maximum cell count of 108, as delineated in Table~\ref{tab:tabledataset}. After employing the filtration and encoding methods described in Sections~\ref{subsec:dbfilter}, \ref{subsec:dv_enc}, and \ref{subsec:standard}, we constructed our pretraining corpus based on the type of data. The corpus is bifurcated into two segments:

\noindent\textbf{Dual-Corpus Objectives Datasets.} This segment is arranged according to the following mappings:
\begin{itemize}
    \item NL+ Schema $\leftrightarrow$ DV query
    \item DV query + Schema $\leftrightarrow$ Description
    \item Table $\leftrightarrow$ Description
    \item Question + DV query + Schema + Table $\leftrightarrow$ Answer
\end{itemize}
As shown in Figure~\ref{fig:pre_frame}, the aforementioned four data types are sequentially presented.

\noindent\textbf{MLM Objectives Datasets.} This segment amalgamates NL questions and database schemas from NVbench, DV queries, questions and answers from FeVisQA, and tables with their descriptions from Chart2Text and WikiTableText. These elements are integrated and then utilized to formulate the Masked Language Model (MLM) pretraining tasks. To illustrate this, a sample DV query from NVBench, which has been subjected to masking, is provided in Figure~\ref{fig:pre_frame}.

\subsection{Data Partitioning}
\label{subsec:data_partition}
After preprocessing the data, we proceeded with the partitioning process. Originating from the Spider dataset \cite{yu2018spider}, NVBench features a wide range of domains, including academic, railway, and scholar, which is conducive to cross-domain evaluation. The data from NVBench was divided into training, validation, and testing subsets, constituting 70\%, 10\%, and 20\% of the dataset, respectively, to facilitate this cross-domain assessment. 
Furthermore, considering that FeVisQA utilizes databases from Spider, we maintained consistency with NVBench by applying the same cross-domain partitioning scheme. The partitioning of the data adheres to the original division as specified in Table~\ref{tab:tabledataset}.

\section{Experiments and Results}
\label{sec:exp}
To comprehensively assess our pre-trained architecture and promote further study, we have assembled the Jointly Understanding Text and Data Visualization benchmark. This benchmark encompasses four extensively studied tasks: text-to-vis (Section~\ref{subsec:text2vis}), vis-to-text (Section~\ref{subsec:vis2text}), FeVisQA (Section~\ref{subsec:qa}), and table-to-text (Section~\ref{subsec:table2text}).
We incorporate established datasets pertinent to these tasks. For each task, we delineate the task definition, baselines, evaluation metrics, corresponding results, and case studies. Additionally, we perform ablation studies on the critical design elements.

\vspace{-5pt}
\subsection{Implementation Details}
\label{subsec:imp}
We conducted the pre-training of DataVisT5 over the course of five epochs using four NVIDIA 40GB A40 GPUs. And we standardized the maximum sequence lengths for both the input and output at 512 tokens. Our training regimen adopted a linear warm-up schedule with a 0.1 warm-up rate and set the learning rate to 5e-6. For optimization, we utilized the DeepSpeedCPUAdam optimizer with a weight decay of 0.01. Further enhancing our training efficiency, we implemented DeepSpeed’s ZeRO Stage 2 offloading strategy with mixed precision (FP16) as described in \cite{10.1145/3394486.3406703}.During the fine-tuning phase, the model exhibited significant sensitivity to hyperparameters, notably the learning rate and training epochs. A grid search was executed to determine the optimal parameters, with selection based on the performance metrics from the validation set across all models. Specifically, for multi-task fine-tuning, parameter optimization was informed by the mean performance across four tasks.


\subsection{Text-to-Vis}
\label{subsec:text2vis}
\begin{table*}[th!]
\caption{Comparative evaluation of text-to-vis models and LLMs performance on the cross-domain NVBench test dataset: non-join operations and complete NVBench with join operations. Best results are highlighted in bold.}  
\vspace{-5pt}
\center
\resizebox{\textwidth}{!}{
    \begin{tabular}{lc|cccc|cccc}
\hline
\textbf{Model} & \textbf{Setting} & \multicolumn{4}{c|}{\textbf{NVBench w/o join operation}} & \multicolumn{4}{c}{\textbf{NVBench w/ join operation}} \\
- &  & Vis EM & Axis EM & Data EM & EM & Vis EM & Axis EM & Data EM & EM \\ \hline
Seq2Vis &  & 0.8027 & 0.0000 & 0.0024 & 0.0000 & 0.8342 & 0.0000 & 0.0000 & 0.0000 \\
Transformer &  & 0.8598 & 0.0071 & 0.0646 & 0.0024 & 0.9798 & 0.0021 & 0.0404 & 0.0000 \\
ncNet &  & 0.9311 & 0.2442 & 0.5152 & 0.1465 & --- & --- & --- & --- \\
RGVisNet &  & 0.9701 & 0.5963 & 0.5423 & 0.4675 & --- & --- & --- & --- \\ 
CodeT5+ (220M) & +SFT& 0.9795 & 0.7889 & 0.6239 & 0.6010 & 0.9843 & 0.4065 & 0.3425 & 0.2968 \\
CodeT5+ (770M) & +SFT& 0.9827 & 0.7850 & 0.6696 & 0.6668 & 0.9865 & 0.4024 & 0.3713 & 0.3399 \\
\hline
GPT-4 (5-shot) & +Similarity & 0.9700 & 0.5507 & 0.6425 & 0.4726 & 0.9790 & 0.2755 & 0.3708 & 0.2313 \\
LLama2-7b & +LoRA& 0.9323 & 0.7432 & 0.6203 & 0.6420 & 0.9446 & 0.4281 & 0.3174 & 0.3327 \\
Mistral-7b & +LoRA & 0.9821 & 0.7753 & 0.6649 & 0.6761 & 0.9246 & \textbf{0.4310} & 0.3386 & 0.3374 \\ \hline

DataVisT5 (220M) & +MFT& 0.9827 & \textbf{0.8078} & 0.6680 & 0.6688 & 0.9873 & 0.4123 & 0.3586 & 0.3324 \\ 
DataVisT5 (770M) & +MFT& \textbf{0.9850} & 0.7983 & \textbf{0.6770} & \textbf{0.6833} & \textbf{0.9884} & 0.4112 & \textbf{0.3863} & \textbf{0.3451} \\ \hline
\end{tabular}%
 }
\label{tb:nl2vis}
\vspace{-10pt}
\end{table*}

\noindent\textbf{Defination.} For a natural language query $\{q, S\}$ consisting of a question $q$ that articulates a user's request for a visualization and $S$, the schema of the relevant database $D$, the goal of the text-to-vis task is to generate the appropriate DV query $y$.

\noindent\textbf{Baselines.} We evaluate DataVisT5 against several established baselines for the text-to-vis task. The \emph{Seq2Vis} approach \cite{luo2021synthesizing} interprets the task as machine translation using a Seq2Seq model equipped with attention. The renowned \emph{Transformer} architecture \cite{vaswani2017attention} and the \emph{ncNet} framework \cite{luo2021natural}, which enhances the Transformer with attention-forcing, serve as additional baselines. \emph{RGVisNet} \cite{10.1145/3534678.3539330} utilizes a two-stage process for retrieving DV queries and modifying the prototype. 
For the performance of LLMs, we explored in-context learning through 5-shot similarity prompting with \emph{GPT-4} \cite{openai2024gpt4} and fine-tuning open-source LLMs such as \emph{Llama2-7b} \cite{touvron2023llama2openfoundation} and \emph{Mistral-7b} \cite{jiang2023mistral7b} using LoRA \cite{hu2021loralowrankadaptationlarge}. Using the \emph{CodeT5+} model \cite{wang2023codet5p} as our base architecture, we employ single-task fine-tuning (SFT) without our novel pretraining as a comparison.


\noindent\textbf{Task-specific Corpus.} For the fine-tuning phase of our text-to-vis task, we engaged the NVBench dataset, which was delineated in Section~\ref{subsubsec:nvbench}, originally derived from our pre-training datasets. Contrasting with the pre-training phase, the fine-tuning was conducted with a singular training objective: NL + Schema $\rightarrow$ DV query.

\noindent\textbf{Evaluation Metrics.}
The performance evaluation of our experiment adopts four metrics, analogous to those utilized in \cite{luo2021synthesizing}. Before delving into the specifics, it is necessary to know that each DV query comprises three key elements: the type of visualization (such as bar chart), the configuration of axis (x/y/z), and data with transformation functions (e.g. group). Additionally, let $N$ denote the total count of test samples. The metrics are: (1) Exact Match (EM), which requires a complete match between the predicted and reference DV queries ($EM = N_{equal}/{N}$), (2) Visualization EM (Vis EM), assessing the accuracy of predicted visualization types ($Vis \ EM = N_{vis}/{N}$), (3) Data EM, focused on data points with transformation functions ($Data \ EM = N_{data}/{N}$), and (4) Axis EM, evaluating the congruence of axis components ($Axis \ EM = N_{axis}/{N}$).

\noindent\textbf{Results.}
Results from Table~\ref{tb:nl2vis} show that foundational models like Seq2Vis and Transformer underperform in cross-domain settings. 
Compared to the previous state-of-the-art, RGVisNet, our multi-task finetuned model exhibited a significant 46.15\% improvement in the EM metric on datasets without join operations. Furthermore, it outperformed the in-context learning approach using GPT-4 in scenarios involving join operations, enhancing the EM metric by 44.59\% and 49.2\%. Notably, in these scenarios, where models such as ncNet and RGVisNet have historically struggled, our model achieved an EM of 0.3451. In comparison to high-parameter (7b) open-source LLMs, our 220M DataVisT5 model performed comparably, while the 770M DataVisT5, with only 11\% of the parameters, achieved optimal performance.


\noindent\textbf{Case Study.}
We illustrate the effectiveness of our DataVisT5 model in generating DV queries compared to other baseline models in Table~\ref{tb:case1}. When processing a NL input, the Seq2Vis model fails to recognize essential keywords such as \emph{visualize} and \emph{group by}, and incorrectly identifies the chart type as \emph{scatter}. The Transformer model, although correct in predicting the visualization type, omits significant information. A similar limitation is observed with ncNet, which, despite generating complex DV queries, fails to include the \emph{group by} transformation function. RGVisNet accurately maps the term 'price' to the 'baseprice' column in the rooms table but does not produce the correct aggregate functions, avg and min. The SFT CodeT5+ incorrectly predicts the elements for \emph{group by}. In contrast, our MFT DataVisT5 model accurately constructs the query: "\emph{visualize scatter select avg(rooms.baseprice), min(rooms.baseprice) from rooms group by rooms.decor}", uniquely achieving the correct visualization results.

\begin{figure*}[th!]
\centering
\begin{subfigure}[t]{0.32\textwidth}
    \centering
    \includegraphics[width=\textwidth]{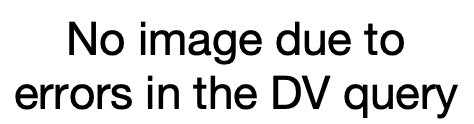}
    \caption{Seq2Vis}
    \label{fig:Seq2Vis_case}
\end{subfigure}
\hfill
\begin{subfigure}[t]{0.32\textwidth}
    \centering
    \includegraphics[width=\textwidth]{figures/error.png}
    \caption{Transformer}
    \label{fig:Transformer_case}
\end{subfigure}
\hfill
\begin{subfigure}[t]{0.32\textwidth}
    \centering
    \includegraphics[width=\textwidth]{figures/error.png}
    \caption{ncNet}
    \label{fig:ncNet_case}
\end{subfigure}
\vspace{5pt}
\begin{subfigure}[t]{0.12\textwidth}
    \centering
    \includegraphics[width=\textwidth]{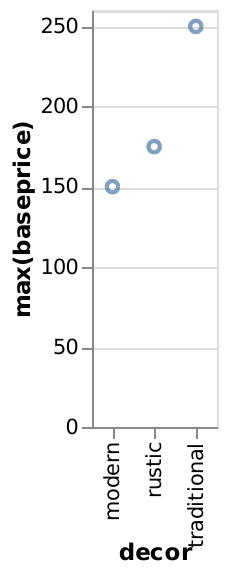}
    \caption{RGVisNet}
    \label{fig:RGVisNet_case}
\end{subfigure}
\hfill
\begin{subfigure}[t]{0.40\textwidth}
    \centering
    \includegraphics[width=\textwidth]{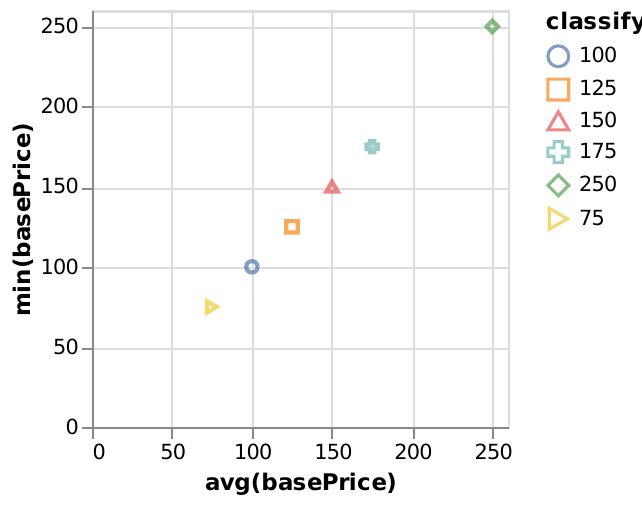}
    \caption{CodeT5+}
    \label{fig:CodeT5_case}
\end{subfigure}
\hfill
\begin{subfigure}[t]{0.43\textwidth}
    \centering
    \includegraphics[width=\textwidth]{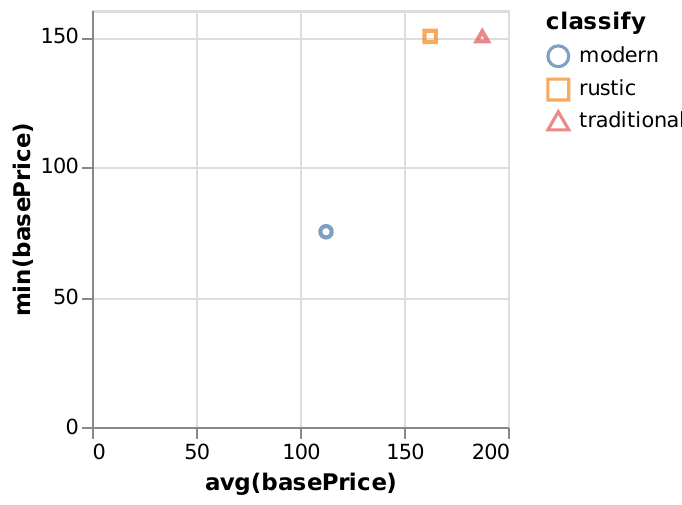}
    \caption{Ours}
    \label{fig:Ours_case}
\end{subfigure}

\label{fig:combined1}
\caption{Visualization formats of DV Query generated by various text-to-vis models}
\end{figure*}

\begin{table*}[htbp]
\centering
\caption{The DV query examples generated by various text-to-vis models from NVBench}
\vspace{-5pt}
\resizebox{\textwidth}{!}{
\begin{tabular}{l|c}
\hline
NL Question & \makecell[l]{Just show the average and minimum price of the rooms in different decor using a scatter.}\\
\hline
Database Schema & \makecell[l]{\texttt{|} inn\_1 \texttt{|} rooms : rooms.roomid, rooms.roomname, rooms.bedtype, rooms.baseprice, rooms.decor}\\
\hline
Ground-truth & \makecell[l]{visualize scatter select avg(rooms.baseprice), min(rooms.baseprice) from rooms group by rooms.decor}\\
\hline
Seq2Vis ($\times$) & \makecell[l]{visualize bar select location, count(company.location) from company group by \\ company.location $\rightarrow$ \textit{Figure~\ref{fig:Seq2Vis_case}}}\\
Transformer ($\times$) & \makecell[l]{visualize scatter select addresses.address\_id, election.vote\_percent from $\rightarrow$ \textit{Figure~\ref{fig:Transformer_case}}}\\
ncNet ($\times$) & \makecell[l]{visualize scatter select rooms.name, rooms.employee\_id from rooms where \\ rooms.first\_name like '\%s\%' $\rightarrow$ \textit{Figure~\ref{fig:ncNet_case}}}\\
RGVisNet ($\times$) & \makecell[l]{visualize scatter select max(rooms.baseprice), rooms.decor from rooms $\rightarrow$ \textit{Figure~\ref{fig:RGVisNet_case}}}\\
CodeT5+ ($\times$) & \makecell[l]{visualize scatter select avg(rooms.baseprice), min(rooms.baseprice) from rooms \\ group by rooms.baseprice $\rightarrow$ \textit{Figure~\ref{fig:CodeT5_case}}}\\
Ours (\checkmark) & \makecell[l]{visualize scatter select avg(rooms.baseprice), min(rooms.baseprice) from rooms \\ group by rooms.decor $\rightarrow$ \textit{Figure~\ref{fig:Ours_case}}}\\
\hline
\end{tabular}
}

\label{tb:case1}
\end{table*}

\vspace{-5pt}
\subsection{Vis-to-Text}
\label{subsec:vis2text}
\begin{table*}[t!]
\caption{Comparative performance analysis of models and LLMs for vis-to-text task . Best results are highlighted in bold.}
\vspace{-5pt}
\resizebox{\textwidth}{!}{%
\begin{tabular}{lcccccccc}
\hline
  \textbf{Method} &\textbf{Setting}&
  BLEU-1 &
  BLEU-2 &
  BLEU-4 &
  ROUGE-1&
  ROUGE-2&
  ROUGE-L&
  METEOR \\ \hline

  Seq2Seq  &&
  0.2766 &
  0.1520 &
  0.0296 &
  0.3571 &
  0.1343 &
  0.2893 &
  0.2528 \\

  Transformer  &&
  0.2825 &
  0.1635 &
  0.0345 &
  0.3634 &
  0.1476 &
  0.2958 &
  0.2755 \\
  BART &+SFT& 0.4301& 0.2892& 0.1009& 0.4721& 0.2209& 0.3647&0.4586\\
  CodeT5+(220M) &+SFT& 0.4431& 0.3060& 0.1236& 0.4873& 0.2403& 0.3770&0.4872\\
  CodeT5+(770M) &+SFT& 0.4518& 0.3154& 0.1278& 0.4898& 0.2431& 0.3928&0.4965\\
  \hline
 GPT-4 (0-shot)& & 0.3843& 0.2210& 0.0387& 0.4180& 0.1527& 0.2925&0.4350\\
 LLama2-7b& +LoRA& 0.3029& 0.1520& 0.0314& 0.3581& 0.1055& 0.2733&0.3028\\
 Mistral-7b& +LoRA& 0.3512& 0.2431& 0.0897& 0.4402& 0.2158& 0.3549&0.3925\\ \hline
  DataVisT5 (220M)&+MFT & 0.4584& \textbf{0.3160}& 0.1245& \textbf{0.5000}& 0.2437& 0.3978&\textbf{0.4986}\\ 
  DataVisT5 (770M) & +MFT &
  \textbf{0.4566}&
  0.3155&
  \textbf{0.1332}&
  0.4974&
  \textbf{0.2460}&
  \textbf{0.3986}&
  0.4851\\ 
\hline
\end{tabular}
}
\label{tb:vis_to_text}
\vspace{-10pt}
\end{table*}

\begin{figure}[htb!]
    \centering
    \includegraphics[width=6cm]{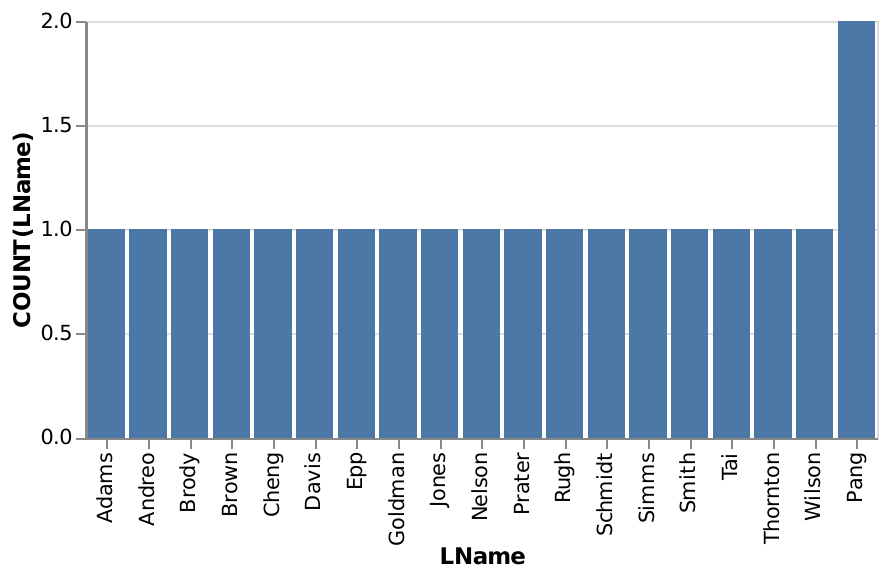}
    \caption{Visualization Chart}
    \label{fig:vis2text_fig1}
    \vspace{-15pt}
\end{figure}

\begin{table*}[th!]
\small
\centering
\caption{The description examples generated by various vis-to-text methods}
\resizebox{\textwidth}{!}{
\begin{tabular}{l|c}
\hline
DV query & \makecell[l]{\textit{Figure~\ref{fig:vis2text_fig1}} $\rightarrow$ visualize bar select student.lname, count(student.lname) from student where stuid not in \\ (select has\_allergy.stuid from has\_allergy join allergy\_type on has\_allergy.allergy =  allergy\_type.allergy \\ where allergy\_type.allergytype  = 'food') group by lname order by count(student.lname) asc }\\
\hline
Database Schema & \makecell[l]{\texttt{|} allergy\_1 \texttt{|} allergy\_type : allergy\_type.allergy, allergy\_type.allergytype \texttt{|} has\_allergy :  has\_allergy.\\stuid, has\_allergy.allergy \texttt{|} student : student.stuid, student.lname, student.fname, student.age, student.sex,\\ student.major, student.advisor, student.city\_code}\\
\hline
Ground-truth & \makecell[l]{List the last name of the students who do not have any food type allergy and count  them in a bar chart,\\ show Y-axis from low to high order.}\\
\hline
Seq2Seq ($\times$) & \makecell[l]{for a bar chart for the the number of the that have the that have the , and a bar chart,  and a bar chart.}\\
\hline
Transformer ($\times$) & \makecell[l]{Find the last names that some last name when that are not steered by any last name as well using a \\ bar chart , and rank by the number of last name in asc .}\\
\hline
BART ($\times$) & \makecell[l]{A bar chart for finding the number of the names of all students who do not have any  allergy with \\ the allergy type "Food", and could you display in ascending by the y-axis?}\\
\hline
CodeT5+ ($\times$) & \makecell[l]{Find the number of students who do not have any allergy type for food in each lname with a bar chart.}\\
\hline
Ours (\checkmark) & \makecell[l]{Give the number of students who do not have any allergy for food in each last name, show by the \\ y-axis from low to high with a bar chart.}\\
\hline
\end{tabular}
}
\label{tb:case2}
\vspace{-10pt}
\end{table*}

\noindent\textbf{Definition.} When provided with a DV query $q$ and a database $D$ that includes a schema $S$, the vis-to-text task is focused on creating an intelligible textual description $z$ that explains the DV query within database schema.

\noindent\textbf{Baselines.} For our evaluation, we selected several established models and LLMs: an enhanced \emph{Seq2Seq} model, which incorporates an attention mechanism as described by \cite{vaswani2017attention} to improve its interaction between the encoder and decoder; the vanilla \emph{Transformer} model as introduced in the context of text-to-vis tasks; \emph{BART} \cite{lewis2019bart}, a transformer-based model that combines bidirectional encoding with auto-regressive decoding; \emph{CodeT5+}, our base architecture; \emph{GPT-4} in a zero-shot setting; and \emph{Llama2-7b} and \emph{Mistral-7b}, both with LoRA fine-tuning.

\noindent\textbf{Task-specific Corpus.} The unidirectional training target for the vis-to-text task was structured as DV query + Schema $\rightarrow$ Description. We employed the NVBench dataset, as referenced in Section~\ref{subsubsec:nvbench}, analogous to the dataset used for the text-to-vis task. A notable distinction for the vis-to-text task lies in the inherent one-to-many relationship, where a singular DV query may correspond to multiple descriptions. To establish a definitive corpus for subsequent fine-tuning and evaluation, we selected a single representative description from the multiples.

\noindent\textbf{Evaluation Metrics.}
To assess the quality of the generated textual descriptions, we employed three metrics: BLEU \cite{papineni2002bleu}, ROUGE \cite{lin2004rouge}, and METEOR \cite{banerjee2005meteor}. (1) BLEU measures the precision of $N$-gram overlaps with reference texts, modified by a brevity penalty. (2) In contrast, ROUGE emphasizes recall, assessing the extent of $n$-gram overlap. (3) METEOR surpasses BLEU in mimicking human judgement by considering exact matches, stemming, synonyms, and penalizing for word order differences. Specifically, we report BLEU scores for unigram, bigram, and four-gram levels (BLEU-1, BLEU-2, BLEU-4), and ROUGE F1 scores for unigrams (ROUGE-1), bigrams (ROUGE-2), and longest common subsequences (ROUGE-L).

\noindent\textbf{Results.} As detailed in Table~\ref{tb:vis_to_text}, the traditional Seq2Seq and Transformer models significantly underperform compared to other models, limited by their parameter size. Although GPT-4 outperforms traditional models in a zero-shot setting, the SFT BART, benefiting from its structure that combines context awareness with autoregressive features, shows superior performance. Moreover, LoRA fine-tuned open-source LLMs Llama2-7b and Mistral-7b, even with larger parameters, do not perform as well as BART, which has significantly fewer parameters, in the vis-to-text task. Despite our base architecture CodeT5+, enhanced through single-task fine-tuning, showing competitive performance, our proposed DataVisT5 in both 220M and 770M configurations achieves the best performance.

\noindent\textbf{Case Study.}
In the comparative analysis presented in Table~\ref{tb:case2}, the Seq2Seq model produces outputs that significantly deviate from the ground truth, indicating a disjointed understanding. The Transformer model, while capturing the basic structure of a bar chart and its ascending order, uses imprecise language that muddles the details. The SFT BART model makes progress by accurately suggesting a bar chart in ascending order but is hampered by suboptimal phrasing. The SFT CodeT5+ model, although closely aligned with the ground truth, fails to grasp the significance of the term \emph{lname} in the visualization context. In stark contrast, our DataVisT5 model, powered by a 770M parameter architecture and enhanced through MFT, excels by providing a concise and clear directive that adeptly delineates the required bar chart with an ascending Y-axis, categorizing students by last name who are without food allergies, thus closely mirroring the ground truth.

\subsection{FeVisQA}
\label{subsec:qa}
\begin{table*}[t!]
\caption{Comparative performance analysis for FeVisQA and table-to-text tasks highlighted by top metric scores.}
\resizebox{\textwidth}{!}{%
\begin{tabular}{lc|cccc|cccc}
\hline
\textbf{Method} & \textbf{Setting} & \multicolumn{4}{c|}{\textbf{FeVisQA}} & \multicolumn{4}{c}{\textbf{Table-to-Text}} \\
- &  & BLEU-1 & ROUGE-1 & ROUGE-L & METEOR & BLEU-4 & ROUGE-1 & ROUGE-L & METEOR \\ \hline
Seq2Seq &  & 0.3642 & 0.3755 & 0.3683 & 0.1955 & 0.1575 & 0.4539 & 0.3995 & 0.3324 \\
Transformer &  & 0.2868 & 0.2984 & 0.2903 & 0.1556 & 0.0875 & 0.3838 & 0.3152 & 0.2642 \\
BART &  +SFT& 0.7379 & 0.7391 & 0.7290 & 0.4376 & 0.3824 & 0.6314 & 0.5549 & 0.5845 \\
CodeT5+(220M) &  +SFT& 0.6813 & 0.6801 & 0.6694 & 0.4086 & 0.3814 & 0.6183 & 0.5450 & 0.5844 \\
CodeT5+(770M) &  +SFT& 0.7039 & 0.7032 & 0.6930 & 0.4211 & 0.3848 & 0.6284 & 0.5511 & 0.5946 \\
\hline
 GPT-4 (0-shot)& & 0.1148& 0.1731& 0.1599& 0.2312& 0.1565& 0.4277& 0.3281& 0.4146\\
 LLama2-7b& +LoRA& 0.4214& 0.4336& 0.4223& 0.2582& 0.2010& 0.4988& 0.4523& 0.3923\\
 Mistral-7b& +LoRA& 0.7404& 0.7671& 0.7574& 0.4251& 0.2003& 0.5002& 0.4538& 0.3948\\ \hline
DataVisT5 (220M) & +MFT& 0.7164 & 0.7158 & 0.7051 & 0.4273 & 0.3822 & 0.6259 & 0.5478 & 0.5926 \\ 
DataVisT5 (770M) & +MFT& \textbf{0.7893} & \textbf{0.7895} & \textbf{0.7788} & \textbf{0.4671} & \textbf{0.4199} & \textbf{0.6520} & \textbf{0.5775} & \textbf{0.6227} \\ \hline
\end{tabular}%
}
\label{tb:result_2}
\vspace{-10pt}
\end{table*}

\noindent\textbf{Definition.} The FeVisQA task is designed to formulate an answer $A$ to a DV-related question $Q$, by leveraging a database $D$ that encompasses a schema $S$ and tables $T$, all in service of elucidating DV concepts.

\noindent\textbf{Baselines.} In addressing the FeVisQA task, we adopted the same ensemble of baseline models previously applied to the vis-to-text task. This ensemble includes an attention-enhanced \emph{Seq2Seq} model, the \emph{Transformer} model, the SFT versions of the base \emph{BART} and \emph{CodeT5+} models, along with a zero-shot \emph{GPT-4}, and LoRA fine-tuned \emph{Llama2-7b} and \emph{Mistral-7b}.

\noindent\textbf{Task-specific Corpus.} The FeVisQA task necessitated the formulation of a unidirectional training objective, structured as: Question + DV query + Schema + Table → Answer. We utilized the FeVisQA dataset for this purpose, which is elaborated upon in Section~\ref{subsubsec:fevisqa} and originates from pre-training datasets.

\noindent\textbf{Results.} From Table~\ref{tb:result_2} In the FeVisQA task, the MFT DataVisT5 model with 770M parameters outperforms competitors across all metrics. Compared to SFT CodeT5+ with an identical parameter setting of 770M, DataVisT5 exhibited a significant 10.92\% increase in the METEOR score post fine-tuning, underscoring its remarkable proficiency in answering free-form questions.
This enhanced performance can be attributed to the integration of textual information and DV knowledge during the DataVisT5 pre-training phase, which effectively facilitates the model's understanding of the complex cross-domain relationship between text and DV.

\begin{figure*}[htb!]
\centering

\begin{subfigure}[b]{0.48\textwidth}
    \centering
    \includegraphics[width=\textwidth]{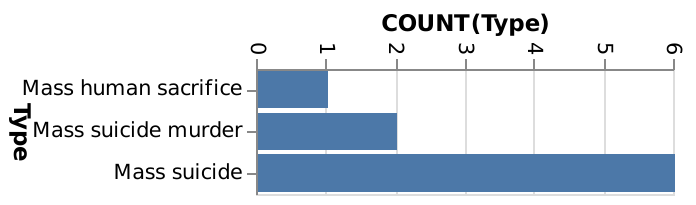}
    \caption{Visualization Chart}
    \label{fig:qa_case1}
\end{subfigure}
\hfill 
\begin{subfigure}[b]{0.4\textwidth}
    \centering
    \includegraphics[width=\textwidth]{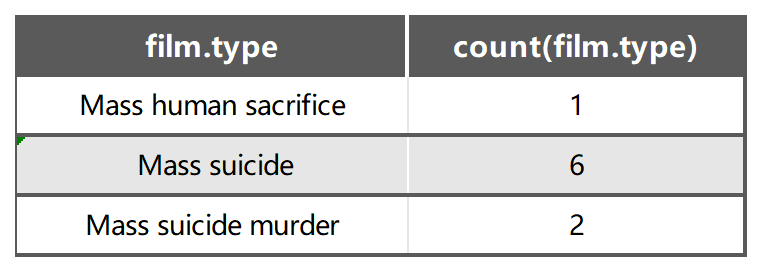}
    \caption{Table}
    \label{fig:qa_case2}
\end{subfigure}

\label{fig:combined}
\caption{Visualization formats of DV Knowledge used in FeVisQA case study}
\vspace{-10pt}
\end{figure*}

\begin{table*}[th!]
\small
\centering
\caption{Sequence formats of DV Knowledge used in FeVisQA case study}
\resizebox{\textwidth}{!}{
\begin{tabular}{l|c}
\hline
DV query & \makecell[l]{\textit{Figure~\ref{fig:qa_case1}} $\rightarrow$ visualize bar select film.type, count(film.type) from film join film\_market\_estimation \\ on film.film\_id = film\_market\_estimation.film\_id group by type order by type asc}\\
\hline
Table & \makecell[l]{\textit{Figure~\ref{fig:qa_case2}} $\rightarrow$ \texttt{|} col : film.type \texttt{|} count(film.type) row 1 : Mass human sacrifice \texttt{|} 1 row 2 : Mass suicide \\ \texttt{|} 6 row 3 : Mass suicide murder \texttt{|} 2}\\
\hline
Database Schema & \makecell[l]{\texttt{|} film\_rank \texttt{|} film : film.film\_id, film.title, film.studio, film.director, film.gross\_in\_dollar \\ \texttt{|} film\_market\_estimation : film\_market\_estimation.estimation\_id,  film\_market\_estimation.low\_estimate, \\ film\_market\_estimation.high\_estimate, film\_market\_estimation.film\_id, film\_market\_estimation.type, \\ film\_market\_estimation.market\_id, film\_market\_estimation.year}\\
\hline
\end{tabular}
}
\label{tb:case3_0}
\end{table*}

\begin{table*}[th!]
\centering
\caption{The answer examples generated by various FeVisQA methods}
\vspace{-5pt}
\resizebox{\textwidth}{!}{
\begin{tabular}{l|c|c|c|c|c|c}
\hline
DV Question & Ground-truth & Seq2Seq & Transformer & BART & CodeT5+ & Ours \\ \hline
\makecell[l]{Is any equal value of y-axis in the chart?} & No & Yes ($\times$) & Yes ($\times$) & Yes ($\times$) & No (\checkmark) & No (\checkmark) \\ \hline
\makecell[l]{How many parts are there in the chart?} & 3 & 5 ($\times$) & 4 ($\times$) & 3 (\checkmark) & 3 (\checkmark) & 3 (\checkmark) \\ \hline
\makecell[l]{What is the value of the smallest part in the chart?} & 1 & 2 ($\times$) & 1 (\checkmark) & 1 (\checkmark) & 1 (\checkmark) & 1 (\checkmark) \\ \hline
\makecell[l]{What is the total number of count(film.type)?} & 9 & 11 ($\times$) & 12 ($\times$) & 15 ($\times$) & 10 ($\times$) & 9 (\checkmark) \\ \hline
\end{tabular}
}
\label{tb:case3}
\end{table*}
\noindent\textbf{Case Study.}
Upon reviewing the outcomes documented in Table~\ref{tb:case3}, we observe that the Seq2Seq, Transformer, and SFT BART models exhibit various discrepancies from the ground truth. The Seq2Seq model consistently produces incorrect responses, indicating significant misalignment. The Transformer model correctly identifies the smallest chart segment but lacks consistency in other queries. SFT BART correctly identifies the number of chart segments but often overestimates numerical values. While the SFT CodeT5+ model answers most questions correctly, it inaccurately responds to \emph{"What is the total number of count(film.type)?"}. In contrast, our DataVisT5 model is the only one that consistently provides accurate answers across both binary and numerical inquiries. 

\vspace{-5pt}
\subsection{Table-to-Text}
\label{subsec:table2text}
\begin{figure*}[htb!]
\centering
\includegraphics[width=0.8\textwidth]{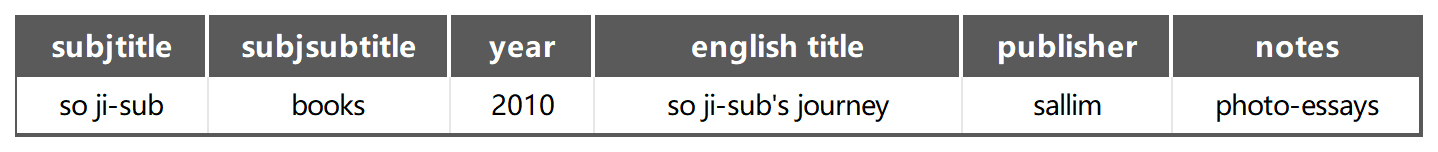}
\caption{Table used in table-to-text case study}
\label{fig:table2text_table}
\vspace{-10pt}
\end{figure*}

\begin{table*}[th!]
\centering
\caption{The description examples generated by various table-to-text methods}
\resizebox{1\linewidth}{!}{
\begin{tabular}{l|c}
\hline
Table& \makecell[l]{\textit{Figure~\ref{fig:table2text_table}} $\rightarrow$ \texttt{|} col : subjtitle \texttt{|} subjsubtitle \texttt{|} year \texttt{|} english title \texttt{|} publisher \texttt{|} notes row 1 : \\ so ji-sub \texttt{|} books \texttt{|} 2010 \texttt{|} so ji-sub's journey \texttt{|} sallim \texttt{|} photo-essays}\\
\hline
Ground-truth & \makecell[l]{Sallim was the publisher of so ji-sub's journey in 2010.}\\  \hline
Seq2Seq ($\times$)& \makecell[l]{the format of was was was was was was.}\\ \hline
Transformer ($\times$)& \makecell[l]{ In movie brand played in 2010.}\\ \hline
BART ($\times$)& \makecell[l]{So ji-sub's journey was published by photo-essays in 2010.}\\ \hline
 CodeT5+ ($\checkmark$)&  \makecell[l]{So ji-sub's journey was published by sallim in 2010.}\\ \hline
Ours ($\checkmark$)& \makecell[l]{Sallim was the publisher of so ji-sub's journey in 2010.}\\
\hline
\end{tabular}
}
\label{tb:case4}
\vspace{-10pt}
\end{table*}

\noindent\textbf{Defination.}
With a table $T$ as the input, the table-to-text task is concentrated on producing a clear, readable narrative $z$ that captures and clarifies the essence of the data within $T$.

\noindent\textbf{Baselines.} Consistent with the previous vis-to-text and FeVisQA tasks, which also focus on text generation modalities, we selected foundational seq-to-seq models for our analysis: the \emph{Seq2Seq} with an attention mechanism, the original \emph{Transformer} model, and fine-tuned versions of the base \emph{BART} and \emph{CodeT5+} models, specifically tailored for single-task applications. Additionally, we included a zero-shot \emph{GPT-4} model and LoRA fine-tuned \emph{Llama2-7b} and \emph{Mistral-7b} in our evaluation.

\noindent\textbf{Task-specific Corpus.} For the table-to-text task, we formulated the unidirectional training target to Table → Description, utilizing a pre-processed pre-training corpus. We amalgamated two publicly accessible datasets, Chart2Text and WikiTableText, which are elaborated upon in Section~\ref{subsubsec:chart2text} and Section~\ref{subsubsec:wikitabletext}.

\noindent\textbf{Results.} As shown in Table~\ref{tb:result_2}, our MFT 770M DataVisT5 model outperforms competing approaches in the table-to-text task, achieving the highest METEOR score of 0.6227. This demonstrates DataVisT5's exceptional ability to generate textual descriptions from tabular data. Foundational models such as Seq2Vis and the Transformer struggle with understanding tables, while the commonly used SFT BART model performs closely to the SFT CodeT5+ (770M) but is still outpaced by DataVisT5. Moreover, the GPT-4 and open-source LLMs also underperform compared to our model. This superior performance is attributed to DataVisT5's integration of textual information and DV knowledge during pre-training.

\begin{table*}[th!]
\caption{Ablation study results: average metric values per task multipled by 100}
\centering
\label{tab:ablation}
\resizebox{\textwidth}{!}{%
\begin{tabular}{llcccccccccc}
\hline
\textbf{Model} & \textbf{Method} & \multicolumn{2}{c}{\textbf{text-to-vis}} & \multicolumn{2}{c}{\textbf{vis-to-text}} & \multicolumn{2}{c}{\textbf{FeVisQA}} & \multicolumn{2}{c}{\textbf{table-to-text}} & \multicolumn{2}{c}{\textbf{Mean}} \\ \hline
DataVisT5 (770M) & MFT & 65.22 &  & 36.18 &  & 70.62 &  & 56.80 &  & 57.21 &  \\ \hline
 & w/o BDC & 64.49 & -0.73$\downarrow$& 36.16 & -0.02$\downarrow$& 69.26 & -1.36$\downarrow$& 55.83 & -0.97$\downarrow$& 56.44 & -0.77$\downarrow$\\
 & w/o up-sampling & 62.95 & -2.27$\downarrow$& 36.41 & +0.23$\uparrow$& 70.69 & +0.07$\uparrow$& 56.34 & -0.46$\downarrow$& 56.60& -0.61$\downarrow$\\
 & w/o MFT& 62.36& -2.87$\downarrow$& 37.12& +0.94$\uparrow$& 67.35& -3.27$\downarrow$& 53.98& -2.82$\downarrow$& 54.93&-2.28$\downarrow$\\
 \hline
 DataVisT5 (770M) & SFT & 65.01 & -0.21$\downarrow$& 36.50& +0.32$\uparrow$& 70.73 & +0.11$\uparrow$& 55.67 & -1.13$\downarrow$& 56.98 &-0.23$\downarrow$\\
CodeT5+ (770M) & SFT & 62.79 & -2.43$\downarrow$& 35.96 & -0.22$\downarrow$& 63.03 & -7.59$\downarrow$& 53.97 & -2.83$\downarrow$& 53.94 & -3.27$\downarrow$\\
T5-large & SFT & 61.34 & -3.88$\downarrow$& 33.58 & -2.60$\downarrow$& 61.90& -8.72$\downarrow$& 52.03 & -4.77$\downarrow$& 52.21 & -5.00$\downarrow$\\ \hline
\end{tabular}%
}
\vspace{-10pt}
\end{table*}

\noindent\textbf{Case Study.}
As detailed in Table~\ref{tb:case4}, the Seq2Seq model's output significantly diverged from the ground truth, producing redundant and irrelevant text without the needed factual content. The Transformer model inaccurately identified the subject as a movie brand rather than a publisher, missing essential details. Although SFT BART correctly identified the publication year and the work's nature, it misattributed the publisher. In contrast, while the SFT CodeT5+ model's responses were semantically close to the ground truth, our model consistently generated descriptions that precisely matched the ground truth.
\vspace{-5pt}
\subsection{Ablation Studies}

We conduct experiments to verify the effectiveness of each critical design in the proposed DataVisT5. Specifically, we establish the MFT DataVisT5 (770M) with all designed components as the baseline. We created variants of DataVisT5 by omitting the BDC objective in the pretraining stage, removing temperature up-sampling during MFT, and evaluating without MFT in a zero-shot setting. Additionally, we compare the use of SFT and MFT, and CodeT5+ versus T5-large as the starting point. From Table~\ref{tab:ablation}, it is evident that removing or replacing designed components results in performance degradation across the mean performance of the four tasks, which indicates the effectiveness of the critical design components in DataVisT5.




\section{Related Work}
\label{sec:related}

\subsection{Pre-training for Data Engineering Tasks}
Pre-trained models have been shown to be effective for language representations and beneficial for downstream tasks by substantial work \cite{pennington2014glove,mccann2017learned,peters2018contextualized,devlin2019bert,raffel2019exploring,liu2019roberta,sun2019ernie,clark2020electra}.
All the success has also driven the development of machine language pretraining, which is in special format of text such as code and sql.
CodeBert~\cite{feng2020codebert} is a bimodal pre-trained model for natural language and programming language in a bert-like architecture, showing that pretraining can improve the performance for code-related tasks. 
TaBert~\cite{yin2020tabert}, TAPAS~\cite{herzig2020tapas} 
 and GraPPa~\cite{yu2020grappa} extend pre-trained models to learn a joint representation of NL text and database tables and demonstrate the effectiveness of semantic parsing tasks. 
Based on pre-trained language models, Rat-SQL \cite{wang2021ratsqlrelationawareschemaencoding} and Proton \cite{wang2022protonprobingschemalinking} enhance text-to-SQL parsing by focusing on schema linking and alignment, whereas StruG \cite{Deng_2021} specifically targets improvements in text-table alignment.

Moreover, the development of domain-adapted pre-trained models, such as CodeT5~\cite{wang2021codet5} for code understanding and generation, MolT5\cite{edwards-etal-2022-translation} for molecule captioning and generation, and BioT5~\cite{pei-etal-2023-biot5} which integrates cross-modal data in the biological domain with chemical structures and linguistic contexts, highlights the importance of specialized training beyond a generic T5 framework. These adaptations emphasize the necessity of domain-specific fine-tuning to effectively capture the contextual nuances inherent in specialized corpora.

\subsection{DV-related Tasks}
Benefiting from the convenience of visualization, various studies related to DV, including text-to-vis, vis-to-text, free-form question answering over DV and table-to-text, have attracted considerable research interest within the community. 
The initial text-to-vis systems were based on predefined rules or templates~\cite{blunschi2011data,zenz2009keywords,shekarpour2015sina,zheng2017natural}. Although efficient, these systems were limited in their ability to handle the linguistic variability of user queries. To overcome these limitations, researchers have turned to neural network-based methods. For example, Data2Vis~\cite{dibia2019data2vis} conceptualizes visualization generation as a sequence translation task, employing an encoder-decoder neural architecture. Similarly, RGVisNet~\cite{10.1145/3534678.3539330} initiates the text-to-vis process by retrieving a relevant query prototype, refining it through a graph neural network (GNN) model, and then adjusting the query to fit the target scenario. Concurrently, vis-to-text has been proposed as a complementary task, with improvements in performance demonstrated through a dual training framework~\cite{song2020vis2text}. Song \textit{et al.}~\cite{song2024} further define the task of free-form question answering over DV and introduce the FeVisQA dataset, aiming to enhance the understanding of data and its visualizations.

Moreover, learning-based approaches have demonstrated exceptional performance in visually data wrangling and analytical tasks.
For instance, Liu \textit{et al.}~\cite{liu2017automatic} and Obeid and Hoque~\cite{obeid-hoque-2020-chart} have successfully translated visual data into textual descriptions and automated natural language summaries for charts using transformer-based architectures, respectively. In a similar vein, Spreafico and Carenini~\cite{10.1145/3399715.3399829} have employed LSTM-based and neural network models to summarize time-series and chart data. Additionally, Kantharaj \textit{et al.}~\cite{kantharaj-etal-2022-chart} have contributed to the evolving benchmark in chart summarization.
Furthermore, Juno\cite{sarkhel2024crossmodalentitymatchingvisually}, a cross-modal entity matching framework, has been developed to contextualize information retrieved from visually rich documents and gather actionable insights, thereby addressing challenges posed by the ad-hoc and often incomplete information in such documents.

\section{Conclusion}
In this study, we propose DataVisT5, a novel PLM specifically designed for DV, which enhances the integration of cross-modal information in DV knowledge and natural language associations. This model introduces a unique mechanism to capture highly relevant database schemas from natural language mentions of tables, effectively unifying and normalizing the encoding of DV knowledge, including DV queries, database schemas, and tables. Our novel hybrid pre-training objectives unravel the complex interplay between DV and textual data, fostering a deeper integration of cross-modal insights. By extending the text-centric T5 architecture to adeptly process cross-modal information, DataVisT5 addresses multiple tasks related to DV with remarkable performance. 
Our extensive experimental results demonstrate that DataVisT5 consistently outperforms SOTA models and even higher-parameter LLMs across a wide range of DV tasks, expanding PLM applications and pushing the boundaries of what is achievable in automated data visualization and interpretation.


\bibliographystyle{IEEEtran}
\bibliography{vis}

\end{document}